\documentclass[11pt]{article}

\usepackage[preprint]{acl}

\usepackage{times}
\usepackage{latexsym}

\usepackage[T1]{fontenc}

\usepackage[utf8]{inputenc}

\usepackage{microtype}

\usepackage{inconsolata}

\usepackage{graphicx}
\usepackage{algorithm}
\usepackage{algorithmic}
\usepackage{amsfonts}
\usepackage{tcolorbox}
\usepackage{pifont}
\tcbuselibrary{breakable} 
\tcbuselibrary{skins} 
\usepackage{enumitem}
\usepackage{xcolor}
\usepackage{colortbl}
\usepackage{amsmath}
\definecolor{lightblue}{RGB}{200,220,240}  
\definecolor{lighterlightblue}{RGB}{235,240,252}  
\newcommand{\bgblue}[1]{\cellcolor{lightblue}#1}
\newcommand{\bglightblue}[1]{\cellcolor{lighterlightblue}#1}

\usepackage{algorithm}
\usepackage{algorithmic}
\usepackage{multirow}
\usepackage{booktabs}
\usepackage{tabularx}
\usepackage{array}
\title{Beyond Token Length: Step Pruner for Efficient and Accurate Reasoning in Large Language Models}

\author{%
    Canhui Wu$^{1,2}$\thanks{Work completed during internship at JD.com}, Qiong Cao$^{2}$\thanks{Corresponding author},  Chang Li$^{2}$,  Zhenfang Wang$^{2}$, Chao Xue$^{2}$ \\ \textbf{Yuwei Fan}$^{1}$, \textbf{Wei Xi}$^{1}$, \textbf{Xiaodong He}$^{2}$ \\
    $^{1}$Xi'an Jiaotong University,  \hspace{0.3em}$^{2}$JD Future Academy \\
    \texttt{wucanhui@stu.xjtu.edu.cn} \quad \texttt{caoqiong1@jd.com}
}

\begin{document}
\maketitle
\begin{abstract}
Large Reasoning Models (LRMs) demonstrate strong performance on complex tasks but often suffer from excessive verbosity, known as "overthinking." Existing solutions via reinforcement learning (RL) typically penalize generated tokens to promote conciseness. However, these methods encounter two challenges: responses with fewer tokens do not always correspond to fewer reasoning steps, and models may develop hacking behavior in later stages of training by discarding reasoning steps to minimize token usage. In this work, we introduce \textbf{Step Pruner (SP)}, an RL framework that steers LRMs toward more efficient reasoning by favoring compact reasoning steps. Our step-aware reward function prioritizes correctness while imposing penalties for redundant steps, and withholds rewards for incorrect responses to prevent the reinforcement of erroneous reasoning. Moreover, we propose a dynamic stopping mechanism to prevent hacking behavior caused by step merging. Extensive experiments across four reasoning benchmarks demonstrate that SP achieves state-of-the-art accuracy while significantly reducing response length. For instance, on AIME24, SP reduces token usage by \textbf{69.7\%}. 


\end{abstract}

%

\section{Introduction}

Reasoning is a fundamental strength of large language models (LLMs), empowering them to address complex tasks that require logical deduction and multi-step inference. Since the advent of Chain-of-Thought (CoT) prompting~\cite{wei2022chain,kojima2022large,yao2023tree}, a wide array of techniques has emerged to further enhance the reasoning capabilities of LLMs. Recent models such as DeepSeek-R1~\cite{guo2025deepseek}, GPT-o1~\cite{jaech2024openai}, Gemini 2.5~\cite{comanici2025gemini}, and QwQ~\cite{team2025qwq} have ushered in a new era of LRMs, which exhibit sophisticated reasoning abilities even in the absence of explicit step-by-step prompts. These LRMs excel in challenging domains, such as mathematical problem-solving and code generation. However, recent research has highlighted a growing inefficiency in the reasoning process of LRMs.

For simple tasks such as ``What is $2+3$?''~\cite{chen2024not}, LRMs can sometimes generate excessively verbose responses, spanning thousands of tokens or falling into cycles of self-doubt, a phenomenon known as \emph{overthinking}~\cite{sui2025stop}. This inefficiency not only leads to increased computational costs~\cite{aytes2025sketch}, but also introduces the risk of compounding errors through unnecessary self-reflection~\cite{cuadron2025danger,su2025between}. In extreme cases, LRMs may even enter infinite reasoning loops, ultimately exhausting their token budgets.

\begin{figure}[t]
    \centering
    \includegraphics[width=\columnwidth]{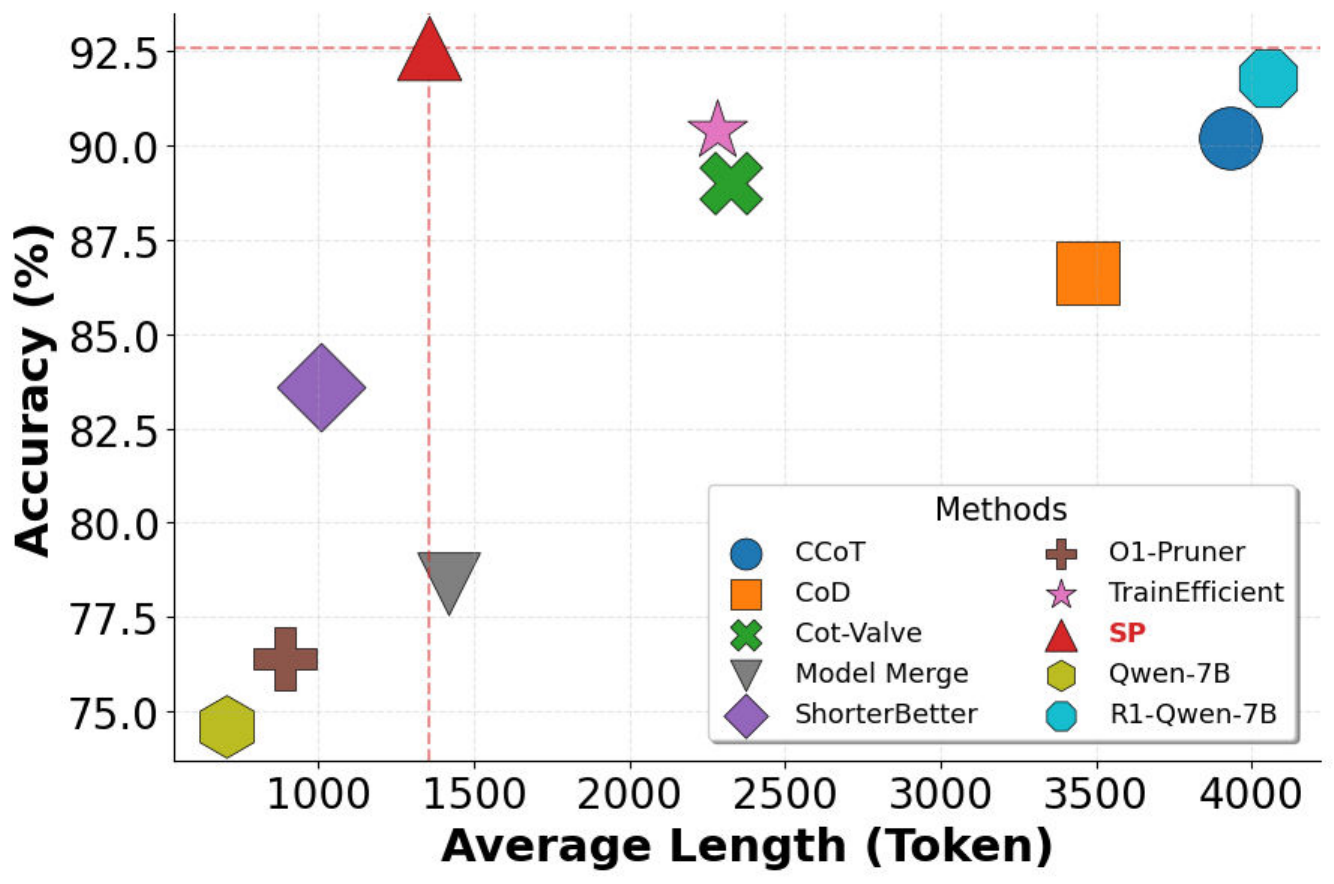}
    \caption{Comparison of Step Pruner with seven baselines on the MATH500 dataset. SP uses only 33\% of the tokens compared to DeepSeek-R1-Distilled-Qwen-7B, and maintains the same level of accuracy.}
    \label{fig:intro}
\end{figure}

To mitigate overthinking, recent studies have proposed several strategies, such as prompting large models to generate concise outputs~\cite{renze2024benefits,xu2025chain} or capping token usage at a predefined limit~\cite{han2024token}. However, these approaches have shown limited effectiveness for LRMs. Supervised fine-tuning (SFT) methods, which use datasets with shorter reasoning, aim to promote brevity. In contrast, RL-based methods offer a more flexible solution by jointly optimizing for both accuracy and output brevity. State-of-the-art RL strategies~\cite{luo2025o1,yi2025shorterbetter,arora2025traininglanguagemodelsreason,hou2025thinkprune} typically penalize token usage while rewarding correctness, thus encouraging models to generate shorter yet accurate answers. For example, o1-pruner~\cite{luo2025o1} introduces an additional reward term by dividing the average length of responses by the current response length, encouraging shorter responses. ShorterBetter~\cite{yi2025shorterbetter} treats the shortest correct response length as optimal and penalizes answers longer than this optimal length.

Despite significant progress, token-based RL faces two main issues: First, fewer tokens do not necessarily mean fewer reasoning steps. Second, in the later stages of training, token-based penalties may incentivize the model to “hack” the reward by outputting only the final answer, omitting the reasoning steps entirely. Inspired by these findings, we propose a novel method, \textbf{Step Pruner (SP)}, which shifts the focus from reducing token count to reducing the number of steps. SP splits the response into distinct steps and penalizes redundant steps, rather than penalizing token units. Our RL-based reward function prioritizes correctness while penalizing redundant steps, thus promoting concise and clear responses. To ensure reliability, the reward for incorrect answers is suppressed, and no step reward is given if the entire response is incorrect. This approach effectively prevents the model from generating shorter but incorrect outputs.
SP also reveals a clear two-phase RL dynamic: 1) redundant steps are first removed, and 2) later, logically related steps begin merging into more compact reasoning units (e.g., collapsing two steps into one paragraph). Finally, we dynamically stop the training by monitoring the model's average output length during the training process. When the model's output no longer shortens, we halt the training. This design enables concise, reliable, and structured reasoning while avoiding model hacking rewards.

Our extensive experiments demonstrate that the SP significantly enhances reasoning efficiency. On multiple benchmarks, SP utilizes only 44\% of the tokens compared to the baselines, while either maintaining or improving accuracy. As shown in Figure~\ref{fig:intro}, SP achieves the best balance between accuracy and token efficiency on the MATH-500 dataset. We also explore alternative segmentation strategies and find that paragraph-based segmentation offers the most effective trade-off between brevity and performance. Additionally, we apply LLM-as-judge for semantic analysis of model outputs before and after training, revealing that the trained model explores fewer redundant paths, resulting in a higher proportion of pivotal reasoning. Our main contributions are summarized as follows:
\begin{enumerate}[leftmargin=*]
    \item We propose \textbf{Step Pruner (SP)}, a reinforcement learning framework that directs LRMs to enhance reasoning efficiency by penalizing redundant steps, and introduce a dynamic stopping mechanism to prevent reward hacking.
    \item We systematically compare various segmentation strategies and find that paragraph-based segmentation strikes the best balance between compression ratio and model performance.
    \item Extensive experiments across four reasoning benchmarks demonstrate that SP achieves the best trade-off between accuracy and response length, outperforming existing baselines.
\end{enumerate}

\section{Related Work}

\subsection{Reasoning in LLMs}
Large language models (LLMs) have demonstrated remarkable capabilities in solving complex reasoning tasks, particularly when guided by carefully designed prompting strategies. A key advancement in this domain is Chain-of-Thought (CoT) prompting~\cite{wei2022chain, kojima2022large, wang2022self}, which breaks down intricate problems into a series of intermediate reasoning steps. To further enhance these abilities, leading LRMs such as OpenAI-O1~\cite{jaech2024openai}, DeepSeek-R1~\cite{guo2025deepseek}, and QwQ~\cite{team2025qwq} integrate advanced techniques like reinforcement learning and multi-stage training. These strategies enable the generation of detailed, multi-step reasoning, achieving state-of-the-art performance in fields such as advanced mathematics and competitive programming.

\subsection{Overthinking in LRMs}
As LRMs advance, the issue of overthinking has become more prominent~\cite{sui2025stop,feng2025efficient}. The reasoning process in these models has grown increasingly verbose, and in some cases, has led to critical issues like infinite repetition. For instance, when answering a simple question like "What is 2 plus 3?"~\cite{chen2024not}, certain reasoning models, particularly smaller ones, may generate reasoning sequences containing thousands of tokens. This verbosity significantly increases the computational cost and latency of reasoning, limiting the practical deployment of these models in computation-sensitive real-world applications~\cite{aytes2025sketch}. In more severe cases, excessive reasoning steps can introduce errors or obscure logical clarity, leading to incorrect answers~\cite{fatemi2025concise}. Beyond a certain threshold, increasing the length of reasoning does not necessarily enhance performance and can even degrade accuracy, as the model may misjudge the complexity of the question and introduce compounded errors~\cite{su2025between}.

\subsection{Efficient Reasoning}
As reasoning in LRMs becomes more verbose, recent works have focused on shortening reasoning while preserving the quality of the generated output. Prompt-based methods, such as CCoT~\cite{renze2024benefits}, modify user prompts by incorporating directives like "Be concise," thereby encouraging the model to generate shorter responses. CoD~\cite{xu2025chain} and Token-budget~\cite{han2024token} introduce a token limit on the model’s output, effectively capping the number of tokens generated and preventing excessively long responses. SFT-based methods, including C3ot~\cite{kang2025c3ot}, Cot-Valve~\cite{ma2025cot} and TokenSkip~\cite{xia2025tokenskip}, focus on creating diverse training datasets with reasoning of varying lengths, with a particular emphasis on shorter reasoning. Fine-tuning the model on these datasets helps it generate more concise reasoning. Merge-based approaches~\cite{wu2025unlocking} reduce reasoning lengths by merging the parameters of a reasoning model with those of a pre-trained instruction model, enhancing efficiency without requiring additional training. Finally, RL-based methods like DAST~\cite{shen2025dast}, O1-pruner~\cite{luo2025o1}, ShorterBetter~\cite{yi2025shorterbetter}, and TrainEfficient~\cite{arora2025traininglanguagemodelsreason} introduce a length penalty during training, rewarding models for generating responses with fewer tokens while maintaining accuracy. These techniques optimize reasoning, making them shorter and more efficient in problem-solving.

\section{Method}
\begin{figure*}[htbp]
    \centering
    \includegraphics[width=\textwidth]{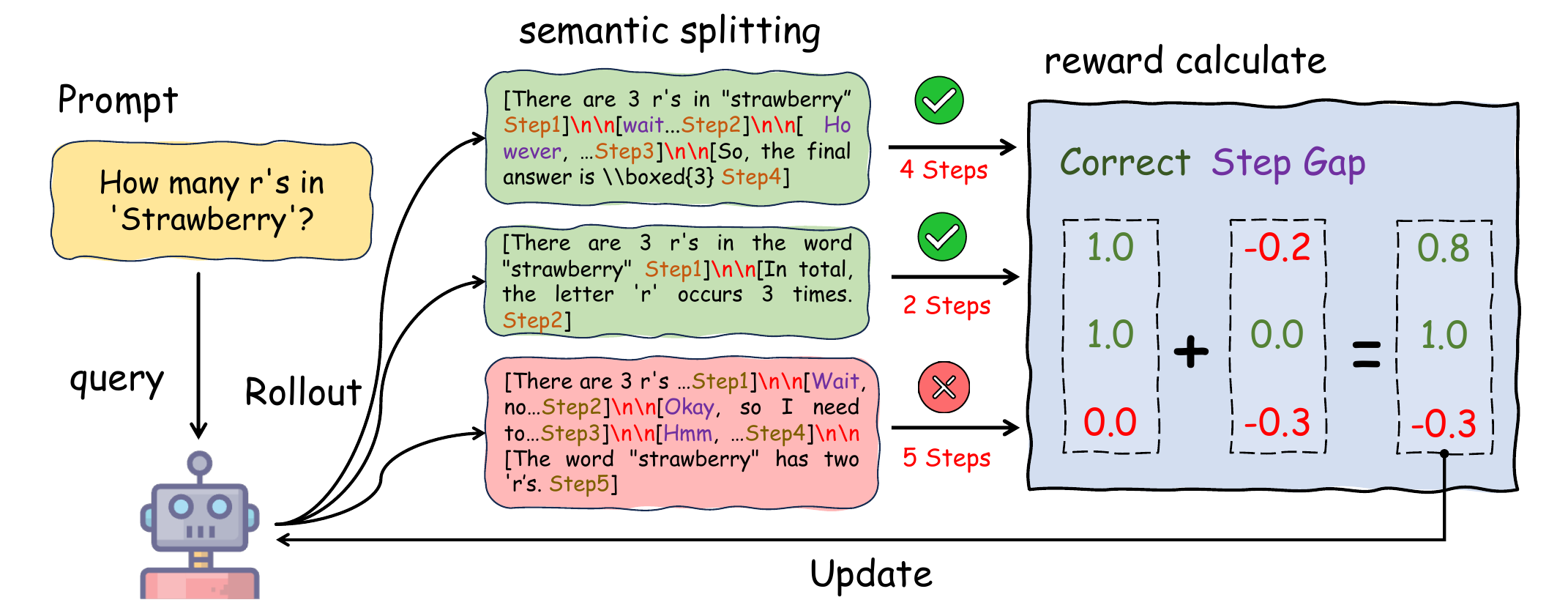}
    \caption{\textbf{Pipeline of Step Pruner}: (1) Prompt LRMs to generate multiple responses; (2) Split each response into logical steps; (3) Calculate a reward score based on both correctness and the number of steps; (4) Update LRMs using GRPO.}
    \label{fig:method}
\end{figure*}

This paper aims to enhance LRM efficiency by introducing the optimal number of steps as a principled metric, rather than directly penalizing response length. Section~3.1 formalizes the objective and notation. Section~3.2 presents our approach for computing optimal steps. Section~3.3 discusses reward configurations balancing correctness and segmentation. Section~3.4 details policy optimization via Group Relative Policy Optimization (GRPO).

\subsection{Problem Setup}

Let $\mathcal{X}$ denote the input space (e.g., prompts), and $\mathcal{Y}$ the output space (e.g., responses). Given $x \in \mathcal{X}$, a large reasoning model $\pi_\theta$ defines a conditional distribution over output sequences. For a response $\mathbf{y} = (y_1, \ldots, y_m) \in \mathcal{Y}$, generation proceeds autoregressively:
\begin{equation}
    \pi_\theta(\mathbf{y}|\mathbf{x}) = \prod_{j=1}^m \pi_\theta(y_j \mid \mathbf{x}, \mathbf{y}_{<j}),
\end{equation}
where $\mathbf{y}_{<j} = (y_1, \ldots, y_{j-1})$.

For each input $x$, let $\mathbf{y}^* \in \mathcal{Y}$ denote the ground-truth answer. The standard objective is to maximize expected correctness:
\begin{equation}
    \max_\theta \ \mathbb{E}_{x \sim \mathcal{D}} \ \mathbb{E}_{\mathbf{y} \sim \pi_\theta(\cdot \mid x)} \left[ R_{\mathrm{acc}}(\mathbf{y}, \mathbf{y}^*) \right],
\end{equation}
where $R_{\mathrm{acc}}$ measures correctness and $\mathcal{D}$ is the data distribution.

In the context of efficient reasoning, we further encourage the model to generate correct answers with minimal verbosity. To achieve this, we introduce a cost function $C(\mathbf{y})$ reflecting response inefficiency. The optimization objective becomes:
\begin{equation}
    \max_\theta \ \mathbb{E}_{x \sim \mathcal{D}} \ \mathbb{E}_{\mathbf{y} \sim \pi_\theta(\cdot \mid x)} \left[ R_{\mathrm{acc}}(\mathbf{y}, \mathbf{y}^*) - \lambda \cdot C(\mathbf{y}) \right],
\end{equation}
where $\lambda > 0$ is a hyperparameter that balances accuracy and efficiency. By designing $C(\mathbf{y})$ appropriately, the model is incentivized to produce responses that are both accurate and concise.

\subsection{Rethinking Length-based Rewards}

Recent RL methods for efficient reasoning often encourage conciseness by penalizing the length of model outputs, assuming that shorter responses correspond to more efficient reasoning. However, this assumption does not always hold: longer responses may contain fewer reasoning steps while presenting a more complete thought process, whereas shorter responses may actually involve more reasoning steps. Moreover, directly penalizing token count can lead to "hacking" phenomena, where the model entirely discards the reasoning process in the later stages of training.

To address these limitations, we propose a step-based reward, which replaces the traditional token penalty with the number of reasoning steps in the response. In practice, we approximate the number of reasoning steps using a simple \textbf{paragraph-based} approach (i.e., separated by \verb|\n\n|), where each paragraph serves as a proxy for a reasoning step~\cite{zhang2025lessons}. The advantage of step-based rewards is that they explicitly indicate the minimal reasoning unit to the model, guiding it to optimize output step by step. Furthermore, step-based training makes it straightforward to identify the appropriate stopping point during training.

\subsection{Reward Function}

We design a reward function that jointly accounts for both response correctness and steps efficiency. The reward consists of two main components: an \emph{accuracy reward} $R_{\mathrm{acc}}$ and a \emph{step reward} $R_{\mathrm{seg}}$.

\noindent
\textbf{Accuracy Reward.}
Given an input $x$ with reference answer $\mathbf{y}^*$, we define the accuracy reward for a generated response $\mathbf{y}$ as a binary indicator:
\begin{equation}
    R_{\mathrm{acc}}(\mathbf{y}, \mathbf{y}^*) = \mathbb{I}[\mathbf{y} = \mathbf{y}^*],
\end{equation}
where $\mathbb{I}[\cdot]$ is the indicator function, returning $1$ for an exact match, and $0$ otherwise.

\noindent
\textbf{Step Reward.}
For each input $x$, we consider $n$ candidate responses $\{\mathbf{y}_1, \ldots, \mathbf{y}_n\}$. Among correct responses ($R_{\mathrm{acc}}(\mathbf{y}_i, \mathbf{y}^*) = 1$), we define the optimal number of steps as:
\begin{equation}
    S^* = \min_{i: R_{\mathrm{acc}}(\mathbf{y}_i, \mathbf{y}^*) = 1} S(\mathbf{y}_i),
\end{equation}
where $S(\mathbf{y})$ is the step count in $\mathbf{y}$. $S^*$ is thus the minimal number of steps needed for correctness.

To avoid rewarding the model for incorrect answers, we mask out all step rewards for incorrect responses. We then distinguish four cases based on correctness and the number of steps:

\begin{itemize}[leftmargin=*,nosep]
    \item[1.] \textbf{Correct, more steps:} If \( R_{\mathrm{acc}}(\mathbf{y}, \mathbf{y}^*) = 1 \) and \( S(\mathbf{y}) > S^* \), we penalize the excess steps. The penalty for excess steps is given by:
    \begin{equation} \label{eq:penalty_correct_more}
    R_{\mathrm{seg}}(\mathbf{y}, S^*) = - (S(\mathbf{y}) - S^*),
    \end{equation}

    \item[2.] \textbf{Correct, fewer steps:} This case is excluded by definition, as the number of steps \( S(\mathbf{y}) \) should always meet or exceed the optimal number \( S^* \) for correctness.

    \item[3.] \textbf{Incorrect, more steps:} If \( R_{\mathrm{acc}}(\mathbf{y}, \mathbf{y}^*) = 0 \) and \( S(\mathbf{y}) > S^* \), we penalize both the error and the excess steps. The penalty is defined as:
    \begin{equation} \label{eq:penalty_incorrect_more}
    R_{\mathrm{seg}}(\mathbf{y}, S^*) = -(S(\mathbf{y}) - S^*),
    \end{equation}

    \item[4.] \textbf{Incorrect, fewer steps:} If \( R_{\mathrm{acc}}(\mathbf{y}, \mathbf{y}^*) = 0 \) and \( S(\mathbf{y}) < S^* \), we penalize the error, but no reward is given for brevity. This case is handled as:
    \begin{equation} \label{eq:penalty_incorrect_fewer}
    R_{\mathrm{seg}}(\mathbf{y}, S^*) = 0,
    \end{equation}
\end{itemize}

Based on the four cases above, the step reward for a response $\mathbf{y}$ can be defined as follows:
\begin{equation}
\small
    R_{\mathrm{seg}}(\mathbf{y}, S^*) = 
    \begin{cases}
        - (S(\mathbf{y}) - S^*) & \text{if } R_{\mathrm{acc}}(\mathbf{y}, \mathbf{y}^*) = 1, \\
        -\max(0, S(\mathbf{y}) - S^*) & \text{if } R_{\mathrm{acc}}(\mathbf{y}, \mathbf{y}^*) = 0, \\
    \end{cases}
\end{equation}
For correct responses, we penalize any excess in the number of steps beyond $S^*$. For incorrect responses, we penalize only if the number of steps exceeds $S^*$, but do not reward responses with fewer steps.

\noindent
\textbf{Final Reward.}
The total reward for a response $\mathbf{y}$ is defined as:
\begin{equation}
    R(\mathbf{y}, \mathbf{y}^*, S^*) = R_{\mathrm{acc}}(\mathbf{y}, \mathbf{y}^*) + \beta \cdot R_{\mathrm{seg}}(\mathbf{y}, S^*),
\end{equation}
where $\beta > 0$ is a hyperparameter that controls the penalty for exceeding the optimal number of steps.

\noindent
\textbf{Handling All-Incorrect Cases.}
To prevent the model from learning to generate incorrect answers, we skip training updates when all candidate responses for input $x$ are incorrect.

\noindent
\textbf{Stopping Criterion.}
During training, we find that step-based reward optimization often progresses in two stages. First, the model steadily shortens its outputs as intended. Then, it begins merging multiple steps into single paragraphs in an attempt to exploit the reward. To avoid this issue, we adopt a simple stopping rule: we stop training once the model's outputs stop shortening. Further details are provided in Section~\ref{appendix:hacking}.

\subsection{Policy Optimization with GRPO}

We optimize the model using the GRPO algorithm~\cite{shao2024deepseekmath}. 
For each input $x$, we sample a group of $n$ candidate responses $\{\mathbf{y}_1, \ldots, \mathbf{y}_n\}$ from the current policy $\pi_\theta$ and compute their total rewards $R(\mathbf{y}_j, \mathbf{y}^*, S^*)$. 
These rewards are then normalized within the group to obtain advantage estimates $\hat{A}_j$. 

We maintain a reference policy $\pi_{\theta_{\mathrm{ref}}}$. 
For each response prefix $\mathbf{y}_{j,1:k}$, we compute likelihoods under both $\pi_\theta$ and $\pi_{\theta_{\mathrm{ref}}}$. 
The GRPO objective encourages higher probability for high-advantage responses while penalizing divergence from the reference policy:
\begin{align}
\small
\mathcal{L}_{\mathrm{GRPO}}(\theta) = 
    -\frac{1}{n} \sum_{j=1}^n \frac{1}{t_j} &\,\sum_{k=1}^{t_j} \Big[ W\big(\hat{A}_j, r_{j,k}, r^{\mathrm{ref}}_{j,k}, \epsilon \big) \notag \\
        &\, - \gamma \cdot \widetilde{\mathrm{KL}} \big( r_{j,k} \,\|\, r^{\mathrm{ref}}_{j,k} \big) 
    \Big],
\end{align}
where $t_j$ is the length of the $j$-th response, $r_{j,k} = \pi_\theta(\mathbf{y}_{j,1:k} \mid x)$, $r^{\mathrm{ref}}_{j,k} = \pi_{\theta_{\mathrm{ref}}}(\mathbf{y}_{j,1:k} \mid x)$, $W(\cdot)$ is a clipped policy improvement term, and $\widetilde{\mathrm{KL}}$ is an approximate KL-divergence penalty. 
Hyperparameters $\epsilon$ and $\gamma$ control clipping and regularization strength.

\section{Experimental Setup}

\subsection{Datasets}
We use DeepScaleR-preview~\cite{luo2025deepscaler} as our training dataset, consisting of 40K math problems (high school to Olympiad level). We evaluate SP on four reasoning benchmarks: AIME24, MATH500, GSM8K, and GPQA. AIME24 contains 30 high-school multi-step problems from the 2024 American Invitational Mathematics Examination. MATH500 is a subset of 500 challenging problems from the MATH dataset~\cite{lightman2023let}. GSM8K~\cite{cobbe2021gsm8k} tests grade-school multi-step arithmetic and algebra. GPQA~\cite{rein2024gpqa} focuses on graduate-level reasoning across diverse fields, requiring true multi-step inference rather than factual retrieval.

\subsection{Model}
We use DeepSeek-R1-Distill‑Qwen‑2.5-7B and DeepSeek-R1-Distill‑Qwen‑2.5-1.5B~\cite{guo2025deepseek} (abbreviated as R1‑Qwen‑7B and R1‑Qwen‑1.5B) as our base models. It is trained by distilling the reasoning traces of DeepSeek-R1 from Qwen-2.5-Math-7B and Qwen-2.5-Math-1.5B~\cite{yang2024qwen25mathtechnicalreportmathematical}.

\subsection{Baselines}

We compare SP with seven recent approaches for efficient reasoning, grouped into four categories:

\noindent
\textbf{Prompt Engineer.}  
Chain of Draft (CoD)~\cite{xu2025chain} and Concise Chain of Thought (CCoT)~\cite{renze2024benefits} promote brevity via prompting, either by guiding models to generate shorter reasoning or by drafting concise initial reasoning with optional elaboration.

\noindent
\textbf{SFT.}
CoT-Valve~\cite{ma2025cot} constructs CoT dataset ranging from short to long by learning a tunable direction in the parameter space, and trains short-thinking models using datasets with shorter CoT.

\noindent
\textbf{Model-merging.}  
Long-to-Short Reasoning~\cite{wu2025unlocking} interpolates between fast and accurate models to balance efficiency and performance.

\noindent
\textbf{RL.} O1-Pruner~\cite{luo2025o1}, ShorterBetter~\cite{yi2025shorterbetter}, and TrainEfficient~\cite{arora2025traininglanguagemodelsreason} use RL to optimize both answer correctness and brevity, where token length is included as a penalty term in the reward function.

\subsection{Evaluations}
We evaluate our models on each benchmark using three metrics:
(1)\textbf{Accuracy}: The proportion of correctly answered questions, directly measuring solution correctness.
(2) \textbf{Response Length}: The average number of tokens per response, indicating the conciseness and efficiency of the model’s reasoning.
(3) \textbf{Accuracy-Efficiency Score} (AES): A composite metric~\cite{luo2025o1} that assigns higher scores to models with greater accuracy and shorter outputs (See the Appendix~\ref{a:aes} for details).

\subsection{Implementation Details}
We utilized a batch size of 128 and generated 4 responses per input with a temperature of 0.9, truncating outputs at 8,000 tokens. The KL-divergence hyperparameter $\gamma$ was set to 0.001, the learning rate was set to 0.001, and in the reward function, the hyperparameter $\beta$ was set to 0.01. For evaluation, we employed the lighteval~\cite{lighteval} toolkit, with a temperature of 0.6.

\section{Experimenta Result}
\subsection{Main Results}

\begin{table*}[tb]
\centering
\renewcommand{\arraystretch}{1.0}  
\setlength{\tabcolsep}{4pt} 
\begin{tabularx}{\textwidth}{l|XXc|XXc|XXc|XXc}

\toprule
\multirow{2}{*}{Method} 
& \multicolumn{3}{c|}{AIME 24} 
& \multicolumn{3}{c|}{GPQA:Diamond} 
& \multicolumn{3}{c|}{MATH500} 
& \multicolumn{3}{c}{GSM8K} \\
 & ACC. & Len. & AES 
 & ACC. & Len. & AES 
 & ACC. & Len. & AES
 & ACC. & Len. & AES \\
\midrule
\multicolumn{12}{c}{\quad \quad \quad \quad \quad \textit{7B Model}} \\
\midrule
Qwen-7B     & 10.0 & 1309 & -- & 30.3 & 724  & --   & 74.6 & 704 & --   & 81.8 & 254 & -- \\
R1-Qwen-7B  & 53.3 & 14839 & 0.00   & 52.0 & 8195  & 0.00   & 91.8 & 4053 & 0.00   & 85.6 & 508 & 0.00 \\
\midrule
CCoT        & \bglightblue{50.0} & 13507 & -0.22  & 51.0 & 8213  & -0.10  & 90.2 & 3931 & -0.06  & 85.2 & 512 & -0.03 \\
CoD         & 43.3 & 13490 & -0.85  & \bglightblue{51.5} & 7761  & 0.00   & 86.6 & 3475 & -0.14  & 84.4 & 463 & 0.02 \\
Cot-Valve   & 46.7 & 10731 & -0.34  & 45.0 & 5386  & -0.33  & 89.0 & 2924 & 0.13   & 84.8 & 439 & 0.09 \\
Model Merge & 26.7 & 5964  & -1.90  & 41.9 & 4425  & -0.51  & 78.2 & 1416 & -0.09   & 81.6 & 636 & -0.49 \\
O1-Pruner   & 26.7 & \bgblue{2574}  & -1.67  & 33.3 & \bgblue{2180}  & -1.06  & 76.4 & \bgblue{993}  & -0.08  & 80.2 & \bgblue{130} & 0.43 \\
ShorterBetter & 43.3 & 4939  & -0.27  & 45.0 & 4326  & -0.20  & 90.4 & \bglightblue{1086} & \bglightblue{0.65}   & 81.6 & \bglightblue{201} & 0.37 \\
TrainEfficient & \bgblue{53.3} & 9798  & \bglightblue{0.34}  & \bgblue{52.0} & 6750 & \bglightblue{0.18}  & \bglightblue{91.6} & 2403 & 0.40   & \bglightblue{85.6} & 227 & \bgblue{0.55} \\
\midrule
\textbf{SP}         & \bglightblue{50.0} & \bglightblue{4502}  & \bgblue{0.39}   & \bglightblue{51.5} & \bglightblue{3925}  & \bgblue{0.47}   & \bgblue{92.0} & 1353 & \bgblue{0.67}   & \bgblue{89.0} & 344 & \bglightblue{0.44} \\
\midrule
\multicolumn{12}{c}{\quad \quad \quad \quad \quad \textit{1.5B Model}} \\
\midrule
Qwen-1.5B  & 0.0 & 5452 & --  & 31.8 & 1174 & --  & 51.8 & 1134 & -- & 64.7 & 296 & -- \\
R1-Qwen-1.5B & 33.3 & 14783 & 0.00 & 33.8  & 9896 & 0.00 & 83.8 & 5794  & 0.00 & 77.6 & 1157 & 0.00 \\
\midrule
CCoT & 26.7 & 16654 & -1.12 & 37.9 & 9065 & 0.45 & \bglightblue{82.4} & 4438 & 0.15 & 70.0 & 419 & 0.15 \\
CoD  & \bgblue{33.3} & 15234 & \bglightblue{-0.03} & \bglightblue{38.9} & 8592 & 0.58 & 79.0 & 3935 & 0.03 & \bglightblue{70.6} & 457 & 0.15 \\
Cot-Valve   & 23.3 & 8736 & -1.09  & 30.8 & 6291  & -0.08  & 75.8 & 3412 & 0.03   & 74.8 & 703 & 0.15 \\
Model Merge & 0.0 & 3977  & -4.27  & 21.9 & 2425  & -1.01  & 12.2 & 3890 & -3.94   & 26.6 & 3428 & -5.25 \\
O1-Pruner   & 23.3 & \bglightblue{3061}  & -0.71  & 32.9 & \bgblue{1887}  & 0.68  & 62.4 & \bgblue{1002}  & -0.45  & 64.7 & \bglightblue{112} & 0.07 \\
ShorterBetter & 23.3 & \bgblue{2908} & -0.70 & 35.4  & \bglightblue{2083} & \bglightblue{0.93} & 68.2  & \bglightblue{1029} & -0.11 & 62.8 & \bgblue{105} & -0.04 \\
TrainEfficient & \bglightblue{30.0} & 10682 & -0.22  & 37.4 & 7114 & 0.60  & 79.2 & 2279 & \bglightblue{0.33} & 70.2 & 134 & \bglightblue{0.41} \\
\midrule
\textbf{SP}         & \bglightblue{30.0} & 5179  & \bgblue{0.15}   & \bgblue{41.9} & 4446  & \bgblue{1.27}   & \bgblue{86.0} & 1779 & \bgblue{0.77}   & \bgblue{80.9} & 563 & \bgblue{0.64} \\
\bottomrule
\end{tabularx}
\caption{Accuracy (\%), response lengths (Len), and AES scores on AIME, GPQA, MATH500, and GSM8K for both 7B and 1.5B models across different baselines. We highlight the \bgblue{best} results and the \bglightblue{second-best} results.}
\label{tab:length-accuracy-aes}
\end{table*}

Table~\ref{tab:length-accuracy-aes} summarizes the performance of SP and seven recent baselines across four benchmarks, for both 7B and 1.5B model sizes.

\noindent\textbf{Accuracy.}
SP achieves consistently strong or best-in-class accuracy across benchmarks and model sizes while markedly shortening responses. On 7B models, SP attains the top performance on MATH500 and GSM8K and remains competitive on GPQA and AIME24. On 1.5B models, SP outperforms all baselines on GPQA: Diamond, MATH500, and GSM8K, and remains highly competitive on AIME24, showing that its advantages hold even at smaller scales.

\noindent\textbf{Efficiency.}
SP produces much shorter outputs than the R1-Qwen baselines at both model sizes. For instance, on AIME24 with the 7B model, SP reduces average output length by 70\%, and by 67\% on MATH500, without compromising accuracy. Similar reductions appear for the 1.5B model. Compared with RL-based baselines such as O1-Pruner and ShorterBetter, SP attains comparable or higher accuracy with only moderately longer responses, offering a more favorable balance between brevity and correctness.

\noindent\textbf{Accuracy-Efficiency Tradeoff.}
The AES metric summarizes the balance between correctness and conciseness. SP attains the highest AES on most benchmarks for both model sizes, reflecting its superior trade-off. Although some baselines like O1-Pruner and Model Merge yield shorter outputs, their substantial accuracy losses reduce overall AES. Prompt-based and SFT-based approaches provide moderate brevity but fall short of SP in combined performance.

\subsection{When to Stop Training to Prevent Hacks}
\label{appendix:hacking}

\begin{figure*}[htb]
    \centering
    \includegraphics[width=\textwidth]{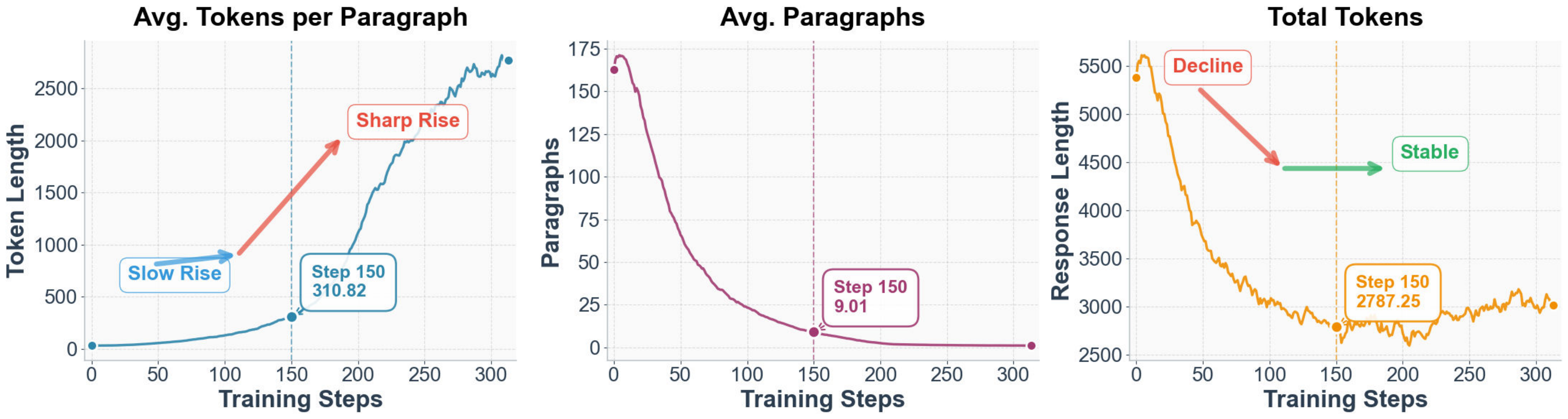}
    \caption{Training curves showing paragraph length, paragraph count, and total response length over training steps.}
    \label{fig:stop}
\end{figure*}

By monitoring various curves during training, we identified the point at which training can be stopped to prevent hacking. As shown in Figure~\ref{fig:stop}, the average number of paragraphs consistently decreased while the average paragraph length increased, and the total token count initially decreased before stabilizing. This pattern reveals two phases in the training process: (1) the paragraph count decreases while paragraph length grows slightly, causing an overall decline in total output length, and (2) paragraph length increases more substantially, the paragraph count levels off, and the total response length stabilizes. Based on these observations, training can be halted once the total output token count ceases to decrease and reaches convergence, effectively preventing hacking.

\subsection{Different Segmentations}

\begin{figure*}[tb]
    \centering
    \includegraphics[width=\textwidth]{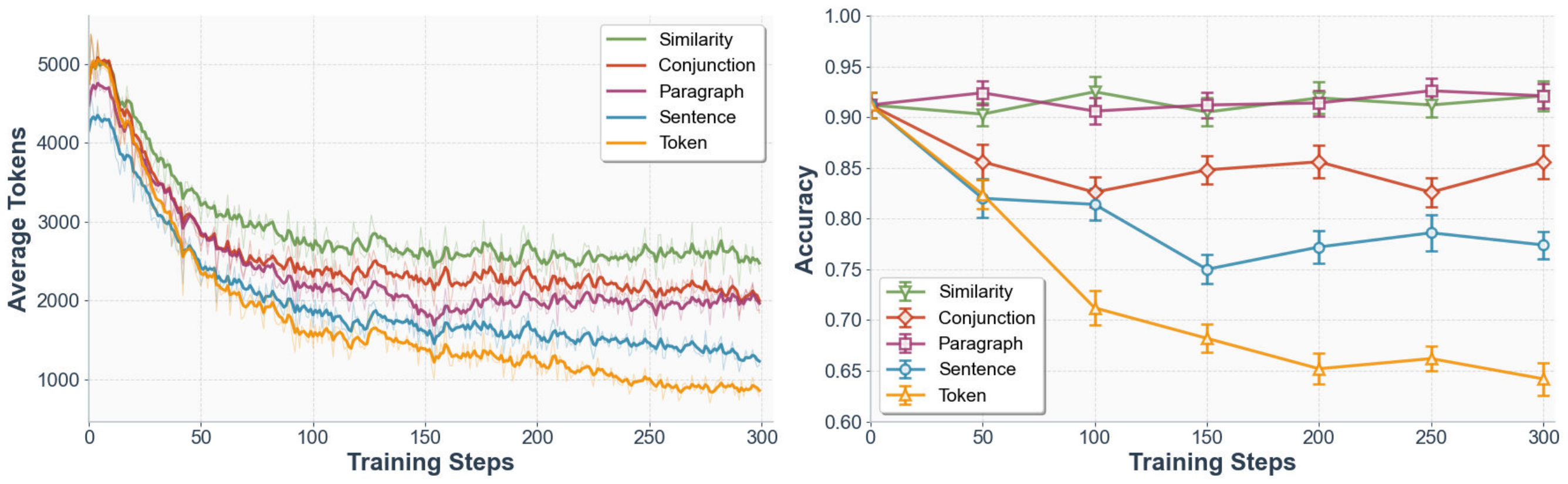}
    \caption{Comparison of different segmentation methods during training: output length (left) and accuracy change (right) on MATH500. The paragraph-based segmentation achieved the best trade-off between accuracy and output length.}
    \label{fig:trend}
\end{figure*}

In addition to basic paragraph-level segmentation, we explore three finer-grained approaches: (1) sentence-level segmentation using NLTK~\cite{bird2006nltk}, (2) conjunction-based segmentation (e.g., "wait", "alternative"; full list in Appendix~\ref{a:conjunctions}), and (3) paragraph merging based on embedding similarity, using \texttt{all-MiniLM-L6-v2} with a cosine similarity threshold of 0.5.

As shown in Figure~\ref{fig:trend}, segmentation based on paragraph similarity results in the smallest output length reduction, stabilizing around 2,500 tokens. Token-level segmentation achieves the greatest reduction (below 1,000 tokens) but significantly impacts accuracy. Paragraph-based segmentation strikes a balance, moderately reducing length while preserving higher accuracy.
In summary, finer-grained segmentation reduces length more effectively but typically lowers accuracy, whereas coarser segmentation better preserves accuracy with less reduction in length.

\subsection{Ablation Studies}

\begin{table}[htb]
\centering

\setlength{\tabcolsep}{3pt}  
\renewcommand{\arraystretch}{0.9}  
\begin{tabularx}{\columnwidth}{l|c|cccc}
\toprule
\textbf{Dataset} & \textbf{Metric} & \textbf{-CR} & \textbf{-COS} & \textbf{-WRM} & \textbf{-SAW} \\
\midrule
\multirow{2}{*}{\textbf{AIME}} 
& Acc. & 0.0 & 36.7 & 33.3 & 36.7 \\
& Len. & 542 & 4379 & 5107 & 5518 \\
\midrule
\multirow{2}{*}{\textbf{MATH}} 
& Acc. & 8.8 & 83.4 & 88.0 & 78.4 \\
& Len. & 348 & 1134 & 1139 & 1224 \\
\midrule
\multirow{2}{*}{\textbf{GSM8K}} 
& Acc. & 22.5 & 85.8 & 86.5 & 84.2 \\
& Len. & 105 & 319 & 344 & 361 \\
\bottomrule
\end{tabularx}
\caption{Ablation study on mathematical tasks. "-" denotes removal of the component. \textbf{CR}: Correct Reward; \textbf{COS}: Correct Optimal Step; \textbf{WRM}: Wrong Response MASK; \textbf{SAW}: Skip All Wrong.}
\label{tab:ablation}
\end{table}

Table~\ref{tab:ablation} summarizes the ablation results across three mathematical datasets. Removing the correctness reward (-CR) causes a clear drop in accuracy, showing that explicit correctness signals are essential for effective policy learning. Prioritizing only shorter answers (-COS) also harms accuracy, highlighting the need to balance brevity with correctness. Without masking wrong answers (-WRM), the model tends to generate shorter yet incorrect outputs, reducing both length quality and accuracy. Eliminating the skip-all-wrong mechanism (-SAW) further degrades performance, indicating the importance of filtering uninformative trajectories. Overall, each component contributes meaningfully to accuracy and stability, and their combination is key to achieving a sound trade-off between correctness and conciseness.

\subsection{Reasoning Analysis}
\begin{figure}[htb]
    \centering
    \includegraphics[width=\columnwidth]{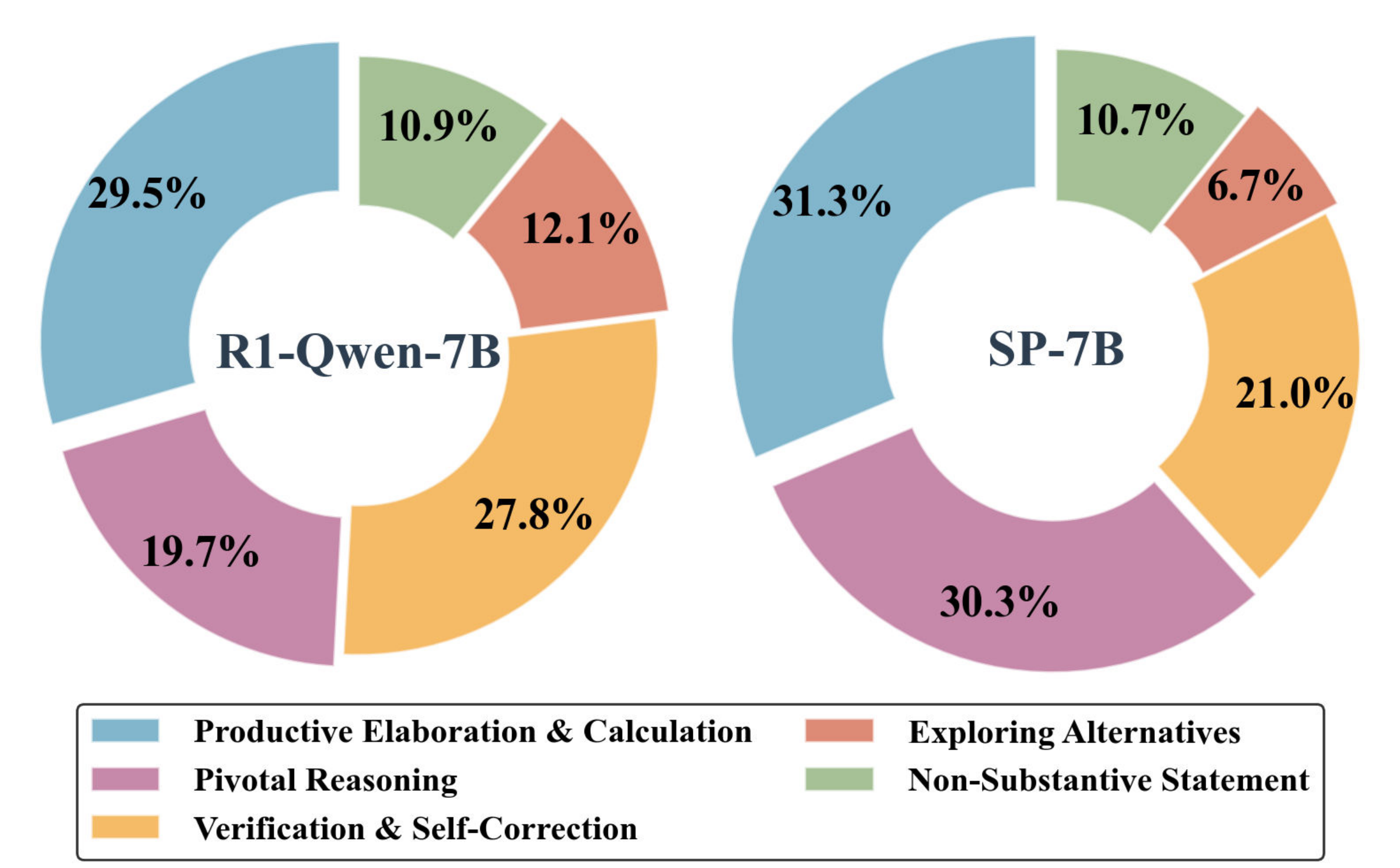}
    \caption{The Semantic Analysis of Reasoning in R1-Qwen-7B and SP-7B.}
    \label{fig:bar}
\end{figure}

In this section, we analyze how SP reshapes the stylistic profile of model‐generated reasonings. We prompted Gemini 2.5 to label each sentence in the R1-Qwen-7B and SP-7B outputs on the MATH500 and AIME24 according to five reasoning categories:

\begin{itemize}[label=\ding{226},leftmargin=*,nosep]
  \item \textbf{Pivotal Reasoning:} Core steps that directly advance the solution.
  \item \textbf{Productive Elaboration \& Calculation:} Supporting explanations or detailed arithmetic.
  \item \textbf{Exploring Alternatives:} Consideration of different methods or perspectives.
  \item \textbf{Verification \& Self-Correction:} In-line checks or corrections of prior steps.
  \item \textbf{Non-Substantive Statements:} Remarks that do not contribute to the logical solution path.
\end{itemize}

Figure~\ref{fig:bar} shows clear shifts in category frequencies. Compared with R1-Qwen-7B, SP-7B produces substantially more Productive Elaboration \& Calculation and Pivotal Reasoning, indicating a stronger focus on detailed computation and essential logical steps. In contrast, Exploring Alternatives and Verification \& Self-Correction decline, suggesting fewer digressions and less frequent self-checking. The share of Non-Substantive Statements remains largely unchanged. Overall, SP-7B encourages more concentrated, substantive, and goal-oriented reasoning.

\section{Conclusion}
This paper introduces Step Pruner (SP), a reinforcement learning framework that enhances the efficiency of LRMs by reducing unnecessary reasoning steps. Unlike methods that directly penalize output length, SP employs a step-based reward mechanism to ensure concise and accurate answers. Experiments on four benchmark tasks show that SP significantly shortens output length while maintaining logical coherence and achieving state-of-the-art accuracy. Further analysis confirms the effectiveness of the step-based reward and each reward component. SP thus provides a simple and robust solution for efficient reasoning in LRMs, offering a practical approach to reduce computational cost while preserving accuracy and logical consistency.

\section*{Limitations}
While Step Pruner effectively reduces redundant reasoning steps and overall response length, it may inadvertently encourage the model to merge distinct logical steps into overly long paragraphs, potentially compromising interpretability and readability. Additionally, the reliance on paragraph-based segmentation may not capture fine-grained reasoning boundaries in all cases, especially for tasks requiring nuanced step differentiation. Finally, SP’s effectiveness is contingent on accurate evaluation of correctness; in domains with ambiguous or subjective answers, reward assignment may be less reliable, limiting generalizability.

\bibliography{main}

\clearpage

\appendix

\section{Appendix}
\label{sec:appendix}

\subsection{Experimental environment}
We trained on the NVIDIA 8*H200 node, with CUDA version 12.7 and CPU model Intel(R) Xeon(R) Platinum 8468V.

\subsection{Accuracy-Efficiency (AE) Score: A Comprehensive Definition}
\label{a:aes}
AES quantifies the trade-off between a model's output length and its accuracy. It evaluates whether a model can reduce the length of its responses without sacrificing accuracy. The AE Score is defined as:

\begin{equation}
\small
\mathrm{AE\text{ } Score} = 
\begin{cases}
\varphi \cdot \Delta\mathrm{Length} + \eta \cdot |\Delta\mathrm{Acc}|, & \text{if } \Delta\mathrm{Acc} \geq 0 \\
\varphi \cdot \Delta\mathrm{Length} - \theta \cdot |\Delta\mathrm{Acc}|, & \text{if } \Delta\mathrm{Acc} < 0
\end{cases}
\end{equation}

In this equation, $\Delta\mathrm{Length}$ and $\Delta\mathrm{Acc}$ represent the percentage changes in output length and accuracy, respectively, between the evaluated model and its baseline. They are defined as follows:

\begin{equation} \small \Delta\mathrm{Length} = \frac{\mathrm{Length}_{\mathrm{baseline}} - \mathrm{Length}_{\mathrm{model}}}{\mathrm{Length}_{\mathrm{baseline}}} \end{equation}

\begin{equation} \small \Delta\mathrm{Acc} = \frac{\mathrm{Acc}_{\mathrm{model}} - \mathrm{Acc}_{\mathrm{baseline}}}{\mathrm{Acc}_{\mathrm{baseline}}} \end{equation}

Here, $\mathrm{Length}_{\mathrm{baseline}}$ and $\mathrm{Acc}_{\mathrm{baseline}}$ denote the output length and accuracy of the baseline model, while $\mathrm{Length}_{\mathrm{model}}$ and $\mathrm{Acc}_{\mathrm{model}}$ refer to those of the evaluated model. A positive $\Delta\mathrm{Length}$ indicates a reduction in output length, and a positive $\Delta\mathrm{Acc}$ indicates an improvement in accuracy.

For our experiments, we set the parameters as in previous work: $\varphi = 1$ (weight for length reduction), $\eta = 3$ (bonus for accuracy improvements), and $\theta = 5$ (penalty for accuracy losses). The asymmetry in the AE Score, with $\theta > \eta$, reflects a practical preference for avoiding accuracy degradation, making the penalty for accuracy drops more significant than the bonus for accuracy gains.

A positive AE Score indicates that the model produces shorter outputs while maintaining or improving accuracy. Conversely, a negative score penalizes any decrease in accuracy.

\subsection{Conjunctions}
\label{a:conjunctions}
The table below lists the conjunctions we use:
\begin{table}[h]
    \centering
    \renewcommand{\arraystretch}{1.2}
    \begin{tabular}{@{}c@{}c@{}}
        \toprule
        \textbf{Conjunctions} \\
        \midrule
        ``wait'', ``alternatively'', ``but'', ``however'',\\ 
        ``alternative'', ``check'', ``double-check'', \\
        ``hmm'', ``okay'', ``maybe''\\
        \bottomrule
    \end{tabular}
\end{table}
\subsection{Prompt}

The prompt used in the Prompt method is shown below.

\begin{tcolorbox}[title=CoD] 
Think step by step, but only keep a minimum draft for each thinking step, with 5 words at most.
\end{tcolorbox}

\begin{tcolorbox}[title=CCoT] 
be concise
\end{tcolorbox}

\subsection{The Hacking Phenomenon in Token-based RL}
\begin{figure*}[tb]
    \centering
    \includegraphics[width=\textwidth]{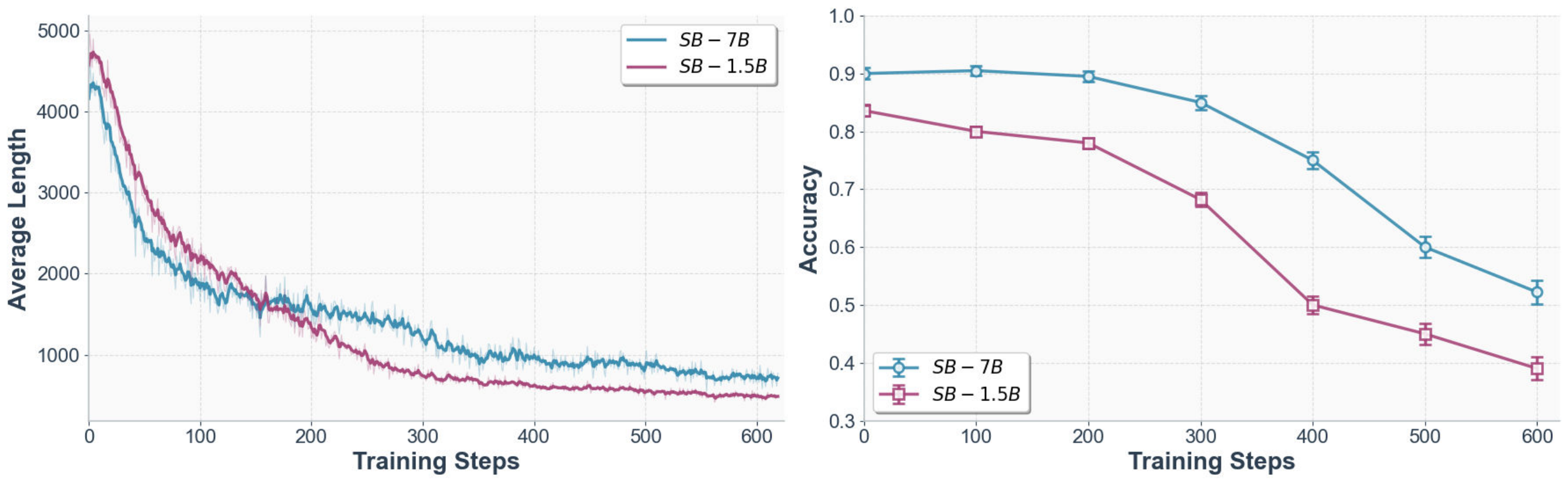}
    \caption{The average output length and accuracy of ShorterBetter during training on the MATH-500 dataset.}
    \label{fig:hacking}
\end{figure*}
The token-based chain-of-thought compression RL method exhibits a hacking reward phenomenon in the later stages of training. Token-based RL methods require continuous supervision, frequent saving of model checkpoints, and regular evaluation to monitor output length and performance. As shown in Figure~\ref{fig:hacking}, ShorterBetter~\cite{yi2025shorterbetter} causes the model to continuously shorten its response length during training. Even after 600 steps, the 7B model can further compress its response, eventually reducing the thinking part to 0 tokens, outputting only \texttt{\textless think\textgreater{}\textbackslash n\textbackslash n\textless /think\textgreater{}}. The figure on the right shows that while the model's performance stabilizes in the first 200 steps, as the output becomes increasingly shorter, the model's performance begins to deteriorate.





\subsection{Why Paragraph Pruning Works}

\begin{figure}[tb]
    \centering
    \includegraphics[width=\columnwidth]{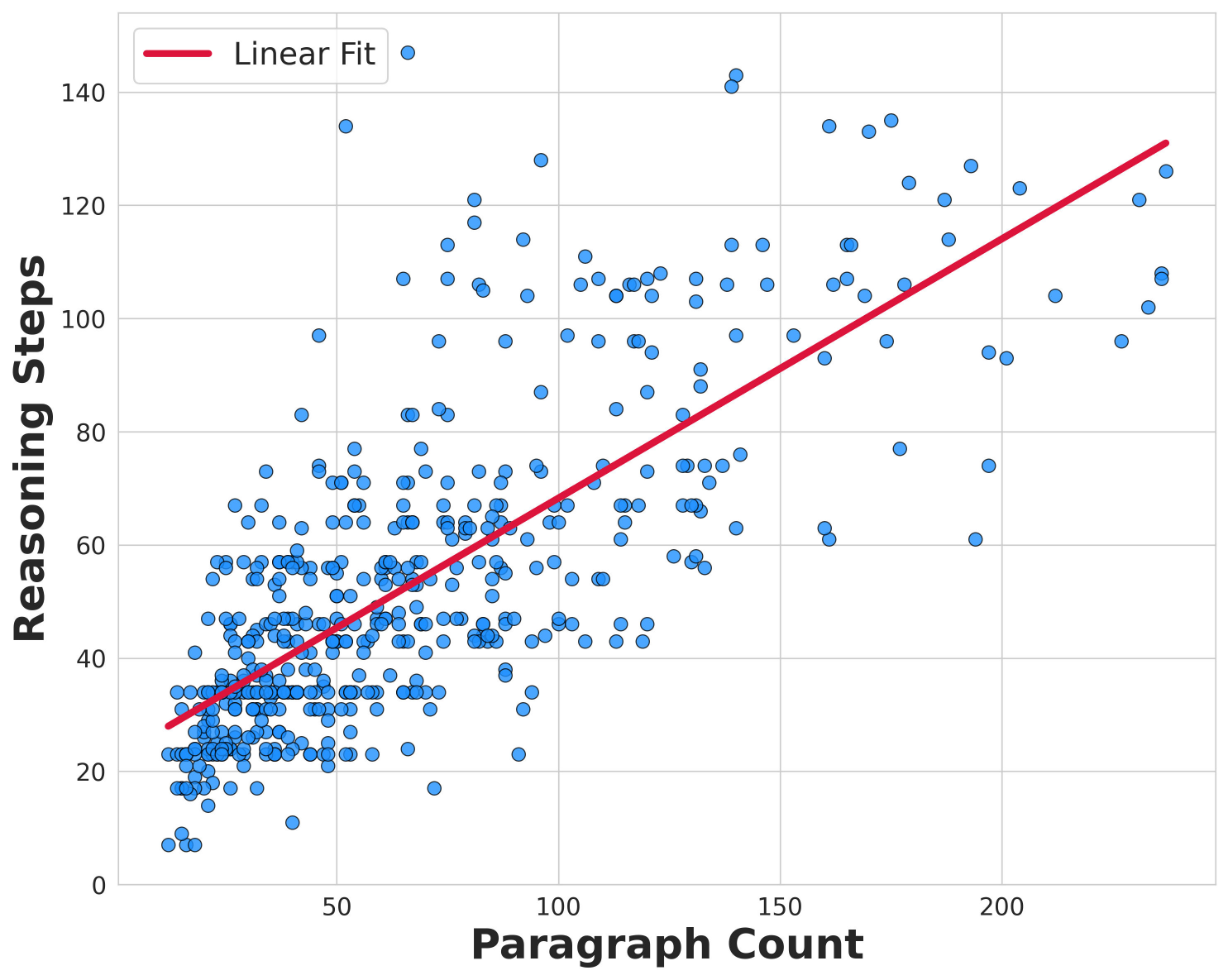}
    \caption{Scatter plot showing the relationship between the number of paragraphs and reasoning steps.}
    \label{fig:linear}
\end{figure}

Although the number of paragraphs is not an exact measure of the number of steps, it serves as a good approximation. To validate this, we sampled 40 questions from three datasets: GPQA, MATH500, and LivecodeBench. For each question, we generated 50 responses, resulting in a total of 2000 answers generated by the DeepSeek-R1-Distill-Qwen-7B model. We then used Gemini 2.5 to decompose the responses into steps, leveraging prompt engineering to estimate the number of steps in each response. Simultaneously, we computed the number of paragraphs for each response by splitting the text at the ``\textbackslash n\textbackslash n'' delimiter.

\begin{tcolorbox}[title=Prompt, before upper={\parskip0pt}, breakable] 
Please analyze the following solution and count the number of reasoning steps. A reasoning step is defined as a distinct logical operation or calculation that moves the solution forward. Paragraphs are split by \textbackslash n\textbackslash n. Even if some paragraphs are short or just verification steps, they should still be counted.

Solution:

\{response\}

Please provide ONLY a number representing the total count of reasoning steps. Put the number in \textbackslash \textbackslash boxed\{\} format.
\end{tcolorbox}

For each question, we calculated the Pearson correlation coefficient between the number of steps and the number of paragraphs across its 50 responses. The resulting correlation coefficient was \textbf{0.661}, demonstrating that the number of paragraphs is a reliable approximation of the number of steps. As shown in Figure~\ref{fig:linear}, the scatter plot displays the relationship between the number of paragraphs and reasoning steps across all responses.

\subsection{Segments or Tokens}

We conducted experiments to assess the effect of adding a token-level length penalty to our training objective. Specifically, we augmented the semantic segmentation loss with an explicit length loss, setting the token length and segment count penalty coefficients to 0.001 and 0.01, respectively. As shown in Table~\ref{tab:segortoken}, incorporating a length penalty led to a substantial reduction in average output length across all benchmarks. However, this brevity came at the cost of a marked decrease in accuracy. These results suggest that directly penalizing output length can harm model performance, likely due to the generation of excessively brief or incomplete solutions.

\begin{table}[htb]
\setlength{\tabcolsep}{4pt}
\centering
\begin{tabular}{l|cccc}
\toprule
 & \textbf{AIME24} & \textbf{GPQA} & \textbf{MATH500} &  \textbf{GSM8K} \\
\midrule
 & Acc. & Acc. & Acc. & Acc. \\
 \midrule
\textbf{P}    & 50.0 & 51.5 & 92.0 & 89.0\\
\textbf{T}    & 40.0 & 48.0 & 64.2 & 66.6 \\
\textbf{P+T}  & 40.0 & 42.4 & 72.0 & 61.8\\
\midrule
 & Len. & Len. & Len. & Len.\\
\midrule
\textbf{P}    & 4502 & 3925 & 1353 & 344\\
\textbf{T}    & 3165 & 3021 & 598 & 92\\
\textbf{P+T}  & 3002 & 2535 & 502 & 109\\
\bottomrule
\end{tabular}
\caption{Impact of different penalties on model response length (Len.) and accuracy (Acc.\%) during training. \textbf{P}: Paragraph-based penalty, \textbf{T}: Token-based penalty.}
\label{tab:segortoken}
\end{table}


\subsection{Ablation experiment on the dataset}

\begin{table*}[t]
\centering
\begin{tabularx}{\textwidth}{l|XXc|XXc|XXc|XXc}

\toprule
\multirow{2}{*}{Method} 
& \multicolumn{3}{c|}{AIME 24} 
& \multicolumn{3}{c|}{GPQA:Diamond} 
& \multicolumn{3}{c|}{MATH500} 
& \multicolumn{3}{c}{GSM8K} \\
 & ACC. & Len. & AES 
 & ACC. & Len. & AES 
 & ACC. & Len. & AES
 & ACC. & Len. & AES \\
\midrule
\multicolumn{12}{c}{\quad \quad \quad \quad \quad \textit{7B Model}} \\
\midrule
Qwen-7B     & 10.0 & 1309 & -- & 30.3 & 724  & --   & 74.6 & 704 & --   & 81.8 & 254 & -- \\
R1-Qwen-7B  & 53.3 & 14839 & 0.00   & 52.0 & 8195  & 0.00   & 91.8 & 4053 & 0.00   & 85.6 & 508 & 0.00 \\
\midrule
\textbf{GRPO}       & 53.3 & 9222  & 0.38   & 50.0 & 6456  & 0.02   & 93.0 & 3009 & 0.30   & 88.8 & 919 & -0.70 \\
\textbf{SP}         & 50.0 & 4502  & 0.39   & 51.5 & 3925  & 0.47   & 92.0 & 1353 & 0.67   & 89.0 & 344 & 0.44 \\
\midrule
\multicolumn{12}{c}{\quad \quad \quad \quad \quad \textit{1.5B Model}} \\
\midrule
Qwen-1.5B  & 0.0 & 5452 & --  & 31.8 & 1174 & --  & 51.8 & 1134 & -- & 64.7 & 296 & -- \\
R1-Qwen-1.5B & 33.3 & 14783 & 0.00 & 33.8  & 9896 & 0.00 & 83.8 & 5794  & 0.00 & 77.6 & 1157 & 0.00 \\
\midrule
\textbf{GRPO}       & 30.0 & 7126  & 0.15   & 37.8 & 5612  & 1.27   & 87.4 & 2706 & 0.77   & 81.8 & 952 & 0.64 \\
\textbf{SP}         & 30.0 & 5179  & 0.15   & 41.9 & 4446  & 1.27   & 86.0 & 1779 & 0.77   & 80.9 & 563 & 0.64 \\
\bottomrule
\end{tabularx}
\caption{Accuracy (\%), response lengths (Len), and AES scores on AIME, GPQA, MATH500, and GSM8K for both 7B and 1.5B models.}
\label{tab:length-accuracy-aes-grpo}
\end{table*}

To eliminate the influence of the dataset, Table~\ref{tab:length-accuracy-aes-grpo} presents the results of standard GRPO training using our dataset. After training with DeepScalar dataset, the model’s performance on several datasets indeed improved. At the same time, SP was able to shorten the response length as much as possible while maintaining accuracy.

\subsection{Ablation Study on Hyperparameters}
\begin{figure*}[htb]
    \centering
    \includegraphics[width=\textwidth]{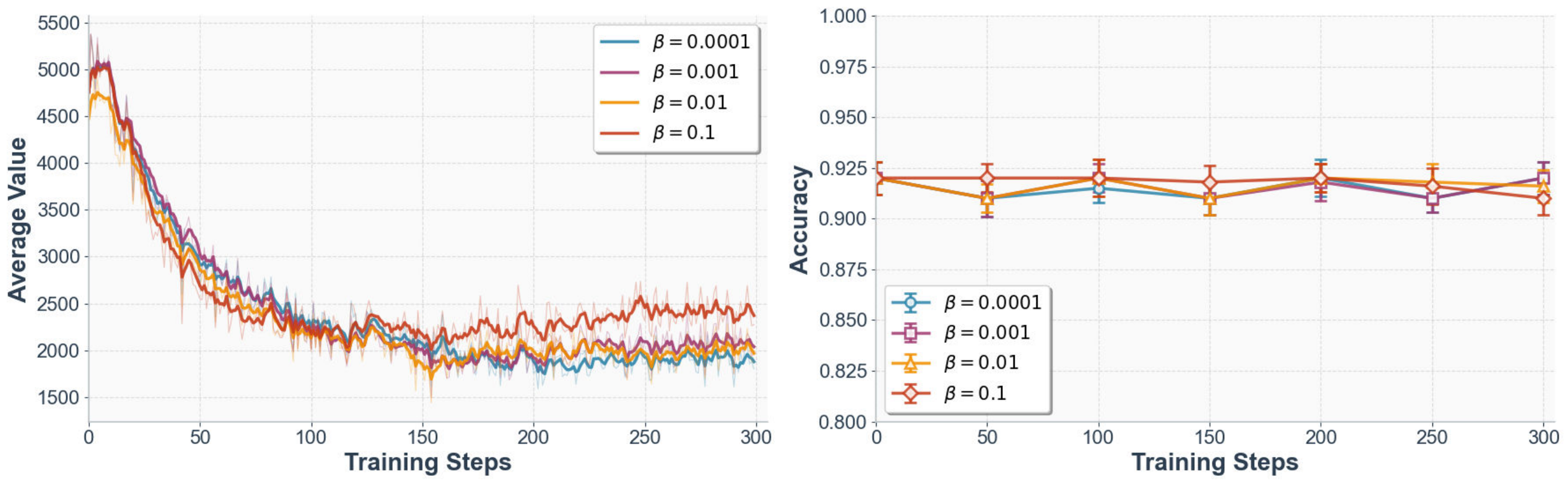}
    \caption{The average output length and accuracy during training for different beta values in the ablation study.}
    \label{fig:beta}
\end{figure*}

As shown in Figure~\ref{fig:beta}, we experimented with multiple settings of the hyperparameter $\beta$. The experimental results indicate that when $\beta$ ranges from 0.01 to 0.0001, the training process remains relatively stable, and the final average output lengths are similar. However, when $\beta$ is set to 0.1, the final output length becomes higher. This is because the increased step reward encourages the model to quickly merge different steps into the same paragraph, resulting in the model learning to exploit the reward early on rather than progressively removing redundant steps during training. Nevertheless, the accuracy on MATH500 remains high across all tested values of $\beta$.

\subsection{Out of Domain Experiment}

\begin{table*}[h]
\centering
\begin{tabularx}{0.7\textwidth}{l|XXc|XXc}
\toprule
\multirow{2}{*}{Method} 
& \multicolumn{3}{c|}{GPQA:Diamond} 
& \multicolumn{3}{c}{Livecodebench} \\
 & ACC. & Len. & AES 
 & ACC. & Len. & AES \\
\midrule
\multicolumn{7}{c}{\quad \quad \quad \quad \quad \textit{7B Model}} \\
\midrule
Qwen-7B     & 30.3 & 724  & --   & 13.8 & 376 & -- \\
R1-Qwen-7B  & 52.0 & 8195  & 0.00   & 34.7 & 13350 & 0.00 \\
\midrule
CCoT        & 51.0 & 8213  & -0.10  & 32.5 & 12521 & -0.25 \\
CoD         & 51.5 & 7761  & 0.00   & 33.2 & 11982 & -0.11 \\
\midrule
Cot-Valve   & 45.0 & 5386  & -0.33  & 29.1 & 13918 & -0.85 \\
\midrule
Model Merge & 41.9 & 4425  & -0.51  & 20.5 & 5517 & -1.46 \\
\midrule
O1-Pruner   & 33.3 & 2180  & -1.06  & 12.1 & 3591 & -2.53 \\
ShorterBetter & 45.0 & 4326  & -0.20  & 33.9 & 7596 & 0.32 \\
TrainEfficient & 52.0 & 6750 & 0.18  & 35.0 & 11224 & 0.19 \\
\midrule
\textbf{SP}         & 51.5 & 3925  & 0.47   & 34.3 & 7209 & 0.40 \\
\midrule
\multicolumn{7}{c}{\quad \quad \quad \quad \quad \textit{1.5B Model}} \\
\midrule
Qwen-1.5B  & 31.8 & 1174 & --  & 4.5 & 842 & -- \\
R1-Qwen-1.5B & 33.8  & 9896 & 0.00 & 14.6 & 16973 & 0.00 \\
\midrule
CCoT & 37.9 & 9065 & 0.45 & 14.2 & 15723 & 0.15 \\
CoD  & 38.9 & 8592 & 0.58 & 14.5 & 14710 & 0.15 \\
\midrule
ShorterBetter & 35.4  & 2083 & 0.93 & 12.3 & 7623 & -0.04 \\
TrainEfficient & 37.4 & 7114 & 0.60  & 16.1 & 11358 & 0.41 \\
\midrule
\textbf{SP}         & 41.9 & 4446  & 1.27   & 16.0 & 9251 & 0.64 \\
\bottomrule
\end{tabularx}
\caption{Accuracy (\%), response lengths (Len), and AES scores on GPQA and Livecodebench for both 7B and 1.5B models across different baselines.}
\label{tab:livecode}
\end{table*}
In addition to GPQA, we also evaluated the Livecodebench. These two datasets belong to the logic and code domains, respectively, which are different from the mathematical domain of the training dataset. The results are presented in Table~\ref{tab:livecode}. Experimental results show that although the shortening effect of SP on GPQA and livecodebench is less significant than that on mathematical datasets, a considerable compression rate is still achieved.

\subsection{Experiments on the Llama Model}

\begin{table*}[tb]
\centering
\renewcommand{\arraystretch}{1.0}  
\setlength{\tabcolsep}{4pt} 
\begin{tabularx}{\textwidth}{l|XXc|XXc|XXc|XXc}

\toprule
\multirow{2}{*}{Method} 
& \multicolumn{3}{c|}{AIME 24} 
& \multicolumn{3}{c|}{GPQA:Diamond} 
& \multicolumn{3}{c|}{MATH500} 
& \multicolumn{3}{c}{GSM8K} \\ 
 & ACC. & Len. & AES 
 & ACC. & Len. & AES 
 & ACC. & Len. & AES
 & ACC. & Len. & AES \\ 
\midrule
\multicolumn{12}{c}{\quad \quad \quad \quad \quad \textit{7B Model}} \\ 
\midrule
Llama-3.1-8B     & 6.7 & 5535 & -- & 33.8 & 3074  & --   & 46.2 & 2131 & --   & 77.6 & 232 & -- \\ 
R1-Llama-8B  & 50.0 & 10628 & 0.0   & 48.9 & 8919  & 0.0   & 87.8 & 3560 & 0.0   & 78.7 & 972 & 0.0 \\ 
\midrule
CCoT        & 43.3 & 11095 & -0.71  & 49.4 & 8230  & 0.11  & 86.4 & 3163 & 0.03  & 70.2 & 516 & -0.07 \\ 
CoD         & 43.3 & 10429 & -0.65  & 48.9 & 7514  & 0.16   & 86.6 & 2854 & 0.13  & 77.4 & 539 & 0.36 \\ 
Model Merge & 13.3 & 18495 & -4.41 & 43.4 & 15882 & -1.34 & 68.8 & 17441 & -4.98   & 85.0 & 19796 & -19.13 \\ 
ShorterBetter & 36.7 & 4291 & -0.73 & 45.2 & 3812 & 0.19 & 78.2 & 1282 & 0.09   & 82.4 & 402 & 0.73 \\ 
TrainEfficient & 46.6 & 5812 & 0.11 & 48.9 & 6283 & 0.30 & 85.2 & 2245 & 0.22   & 84.8 & 591 & 0.62 \\ 
\midrule
SP         & 46.6 & 4917  & \bgblue{0.20}   & 48.9 & 4419  & \bgblue{0.50}   & 84.8 & 1639 & \bgblue{0.37}   & 87.2 & 497 & \bgblue{0.81} \\ 
\bottomrule
\end{tabularx}
\caption{Accuracy (\%), response lengths (Len), and AES scores on AIME, GPQA, MATH500, and GSM8K for 8B Llama-3.1 models across different baselines. We highlight the \bgblue{best} results.}
\label{tab:length-accuracy-aes-llama}
\end{table*}

The results for the Llama-3.1-8B model, as shown in Table~\ref{tab:length-accuracy-aes-llama}, demonstrate that the SP method performs exceptionally well across a variety of datasets. For instance, on the MATH500 dataset, SP achieves an accuracy of 84.8\% with a relatively shorter response length compared to other methods, further highlighting its efficiency in balancing accuracy and response length.

These findings emphasize the robustness of the SP method, confirming its effectiveness not only on the Qwen model but also on the Llama model. The consistent high performance across different base models and datasets underscores the reliability of SP in achieving superior results in a range of settings, especially in terms of accuracy and response length trade-offs.

\subsection{Reasoning Analysis with Different Methods}
\begin{figure}[tb]
    \centering
    \includegraphics[width=\columnwidth]{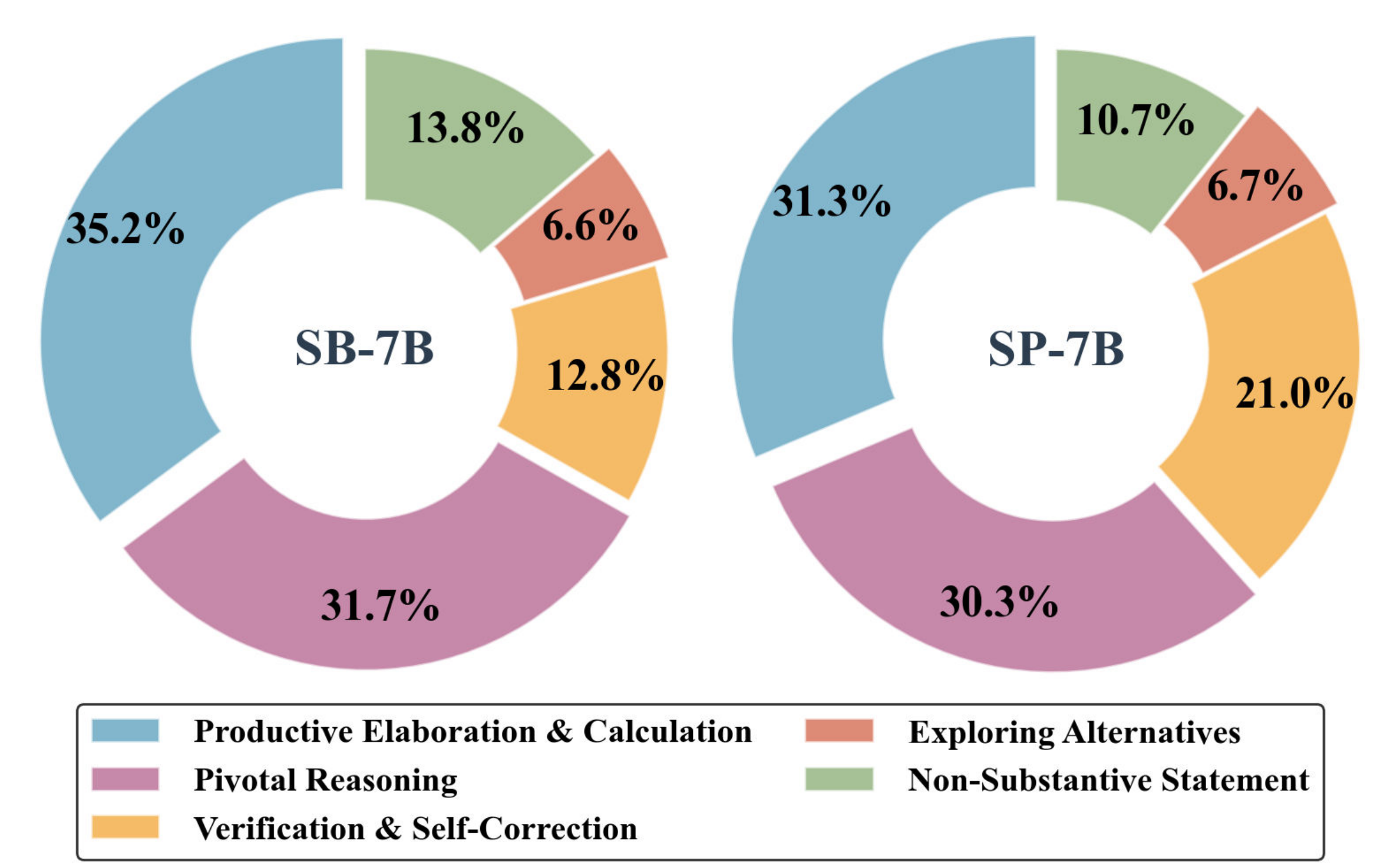}
    \caption{The Semantic Analysis of Reasoning in SB-7B and SP-7B.}
    \label{fig:bar-appendix}
\end{figure}

The SP-7B model demonstrates superior accuracy compared to SB-7B, largely attributable to differences in response structure. We compare our method with ShorterBetter~\cite{yi2025shorterbetter}. SP-7B allocates a higher proportion to "Verification \& Self-Correction" (20.96\% vs. 12.79\%), emphasizing critical evaluation and error-checking, which likely reduces inaccuracies in reasoning. In contrast, SB-7B prioritizes "Productive Elaboration \& Calculation" (35.17\% vs. 31.33\%), focusing on detailed computation but potentially overlooking verification steps. This structural shift in SP-7B promotes more balanced and reliable outputs, as self-correction mitigates over-elaboration risks. While SB-7B excels in exploratory aspects, SP-7B's emphasis on validation enhances overall precision, making it more effective for tasks requiring rigorous accuracy.

\subsection{Answer from Different Methods}
We also present the responses from Shorterbetter and Step Pruner, with the questions taken from AIME24.

\begin{tcolorbox}[colback=yellow!10!white, colframe=red!75!black, title=ShorterBetter (2615 tokens), breakable, float*=tb, width=\textwidth, before upper={\parskip7pt}] 
$<$think$>$

To solve this problem, I need to find the remainder when the product \( \prod_{k=0}^{13}(2 - 2\omega^k + \omega^{2k}) \) is divided by 1000, where \( \omega \neq 1 \) is a 13th root of unity.

First, I recognize that \( \omega \) is a primitive 13th root of unity, so \( \omega^{13} = 1 \). The product is over all 13th roots of unity, including \( \omega^0 = 1 \).

I notice that the expression inside the product can be rewritten as \( 2 - 2\omega^k + \omega^{2k} \). Maybe I can factor this expression or relate it to a polynomial whose roots are the \( \omega^k \).

Let me consider the polynomial \( P(x) = x^{13} - 1 \). Its roots are the 13th roots of unity, including \( \omega^k \) for \( k = 0, 1, ..., 12 \).

I want to relate \( 2 - 2\omega^k + \omega^{2k} \) to \( P(x) \). Maybe I can express it as \( \omega^{2k} - 2\omega^k + 2 \), which resembles a quadratic in \( \omega^k \).

Let me set \( y = \omega^k \). Then the expression becomes \( y^2 - 2y + 2 \). I can write this as \( (y - 1)^2 + 1 \).

So, the product becomes \( \prod_{k=0}^{12} ((\omega^k - 1)^2 + 1) \).

Now, I need to compute this product. I recall that for roots of unity, products involving \( \omega^k - 1 \) can often be related to cyclotomic polynomials or other symmetric products.

Let me consider the polynomial \( Q(x) = \prod_{k=0}^{12} (x - \omega^k) = x^{13} - 1 \).

But I need to compute \( \prod_{k=0}^{12} ((\omega^k - 1)^2 + 1) \). Let me set \( z = \omega^k \), then the term becomes \( (z - 1)^2 + 1 = z^2 - 2z + 2 \).

So, the product is \( \prod_{k=0}^{12} (z^2 - 2z + 2) \) where \( z \) runs over all 13th roots of unity.

This is equivalent to evaluating the product \( \prod_{k=0}^{12} (z^2 - 2z + 2) \) where \( z \) are the roots of \( z^{13} - 1 = 0 \).

I can think of this as evaluating the polynomial \( R(z) = z^2 - 2z + 2 \) at all the 13th roots of unity and taking the product.

There is a formula for such products: \( \prod_{k=0}^{n-1} R(\omega^k) = \prod_{k=0}^{n-1} R(\omega^k) \), which can be related to the resultant or the value of another polynomial.

Alternatively, I can consider that \( \prod_{k=0}^{12} (z - \omega^k) = z^{13} - 1 \). So, if I set \( z = \sqrt{2} e^{i\theta} \), but that might complicate things.

Wait, another approach: The product \( \prod_{k=0}^{12} (z^2 - 2z + 2) \) can be written as \( \prod_{k=0}^{12} ((z - (1 + i))(z - (1 - i))) \).

So, the product becomes \( \prod_{k=0}^{12} (z - (1 + i))(z - (1 - i)) \) evaluated at \( z = \omega^k \).

Therefore, the entire product is \( \prod_{k=0}^{12} (1 + i - \omega^k)(1 - i - \omega^k) \).

This can be written as \( \prod_{k=0}^{12} (1 + i - \omega^k) \times \prod_{k=0}^{12} (1 - i - \omega^k) \).

Each of these products is the value of the polynomial \( P(x) = x^{13} - 1 \) evaluated at \( x = 1 + i \) and \( x = 1 - i \), respectively.

So, \( \prod_{k=0}^{12} (1 + i - \omega^k) = (1 + i)^{13} - 1 \) and similarly \( \prod_{k=0}^{12} (1 - i - \omega^k) = (1 - i)^{13} - 1 \).

Therefore, the original product is \( [(1 + i)^{13} - 1][(1 - i)^{13} - 1] \).

Now, I need to compute \( (1 + i)^{13} \) and \( (1 - i)^{13} \).

First, note that \( 1 + i = \sqrt{2} e^{i\pi/4} \), so \( (1 + i)^{13} = (\sqrt{2})^{13} e^{i13\pi/4} \).

Similarly, \( (1 - i)^{13} = (\sqrt{2})^{13} e^{-i13\pi/4} \).

Compute \( (\sqrt{2})^{13} = (2)^{13/2} = 2^{6} \times 2^{1/2} = 64 \sqrt{2} \).

Now, \( e^{i13\pi/4} = e^{i(3\pi + \pi/4)} = e^{i\pi/4} \times e^{i3\pi} = e^{i\pi/4} \times (-1) = -e^{i\pi/4} \).

Similarly, \( e^{-i13\pi/4} = -e^{-i\pi/4} \).

Therefore, \( (1 + i)^{13} = 64 \sqrt{2} \times (-e^{i\pi/4}) = -64 \sqrt{2} e^{i\pi/4} \).

Similarly, \( (1 - i)^{13} = -64 \sqrt{2} e^{-i\pi/4} \).

Now, compute \( (1 + i)^{13} - 1 = -64 \sqrt{2} e^{i\pi/4} - 1 \).

Similarly, \( (1 - i)^{13} - 1 = -64 \sqrt{2} e^{-i\pi/4} - 1 \).

Multiply these two:

\[
[ -64 \sqrt{2} e^{i\pi/4} - 1 ][ -64 \sqrt{2} e^{-i\pi/4} - 1 ]
\]

Let me compute this product step by step.

First, expand the product:

$= [ (-64\sqrt{2} e^{i\pi/4})(-64\sqrt{2} e^{-i\pi/4}) ] + [ (-64\sqrt{2} e^{i\pi/4})(-1) ] + [ (-1)(-64\sqrt{2} e^{-i\pi/4}) ] + [ (-1)(-1) ]$

Compute each term:

1. First term: $(-64\sqrt{2})(-64\sqrt{2}) \cdot e^{i\pi/4} e^{-i\pi/4} = (64\sqrt{2})^2 \cdot e^{0} = (64^2)(2) \cdot 1 = 4096 \cdot 2 = 8192$

2. Second term: $(-64\sqrt{2})(-1) e^{i\pi/4} = 64\sqrt{2} e^{i\pi/4}$

3. Third term: $(-1)(-64\sqrt{2}) e^{-i\pi/4} = 64\sqrt{2} e^{-i\pi/4}$

4. Fourth term: $(-1)(-1) = 1$

So, adding all terms together:

$8192 + 64\sqrt{2} e^{i\pi/4} + 64\sqrt{2} e^{-i\pi/4} + 1$

Combine the exponential terms:

$64\sqrt{2} (e^{i\pi/4} + e^{-i\pi/4}) = 64\sqrt{2} \cdot 2 \cos(\pi/4) = 64\sqrt{2} \cdot 2 \cdot (\sqrt{2}/2) = 64\sqrt{2} \cdot \sqrt{2} = 64 \cdot 2 = 128$

So, the total becomes:

$8192 + 128 + 1 = 8321$

Therefore, the product is $8321$.

Now, we need the remainder when $8321$ is divided by $1000$.

$8321$ divided by $1000$ is $8$ with a remainder of $321$.

So, the remainder is $321$.

\textbf{Final Answer}
\[
\boxed{321}
\]

$<$/think$>$

Given \(\omega \neq 1\) is a 13th root of unity, we need to find the remainder when the product

\[
\prod_{k=0}^{12}(2 - 2\omega^k + \omega^{2k})
\]

is divided by 1000.

First, we rewrite the expression inside the product as \(2 - 2\omega^k + \omega^{2k}\), which can be expressed as \((\omega^k - 1)^2 + 1\). Thus, the product becomes

\[
\prod_{k=0}^{12} ((\omega^k - 1)^2 + 1).
\]

We recognize that this product can be related to evaluating the polynomial \(P(x) = x^{13} - 1\) at specific points. We rewrite the product as

\[
\prod_{k=0}^{12} (z^2 - 2z + 2)
\]

where \(z\) runs over all 13th roots of unity. This product is equivalent to evaluating the polynomial at \(z = 1 + i\) and \(z = 1 - i\), respectively. Therefore, we need to compute

\[
[(1 + i)^{13} - 1][(1 - i)^{13} - 1].
\]

We find that \((1 + i)^{13} = -64\sqrt{2} e^{i\pi/4}\) and \((1 - i)^{13} = -64\sqrt{2} e^{-i\pi/4}\). Thus, the product becomes

\[
[ -64 \sqrt{2} e^{i\pi/4} - 1 ][ -64 \sqrt{2} e^{-i\pi/4} - 1 ].
\]

Expanding this product, we get:

\[
8192 + 128 + 1 = 8321.
\]

Finally, the remainder when 8321 is divided by 1000 is

\[
\boxed{321}.
\]
\end{tcolorbox} 

\begin{tcolorbox}[colback=yellow!10!white, colframe=blue!75!black, title=SP (2037 tokens), breakable, float*=tb, width=\textwidth, before upper={\parskip7pt}] 
$<$think$>$

Okay, so I have this problem:

Let $\omega \neq 1$ be a $13$th root of unity. I need to find the remainder when the product from $k=0$ to $12$ of $(2 - 2\omega^k + \omega^{2k})$ is divided by $1000$.

Hmm, that seems a bit complicated, but let me try to break it down.

First, I know that $\omega$ is a $13$th root of unity, so $\omega^{13} = 1$. That might be useful.

The product is from $k=0$ to $12$, which is all the roots, so that's the complete set of $13$th roots of unity.

Hmm, okay. So each term in the product is $2 - 2\omega^k + \omega^{2k}$. Maybe I can simplify that expression somehow.

Let me see, $2 - 2\omega^k + \omega^{2k}$... Hmm, perhaps I can write it as a quadratic in $\omega^k$.

Let me set $x = \omega^k$, then the expression becomes $2 - 2x + x^2$.

So, $2 - 2x + x^2$. That's a quadratic, maybe I can factor it or find its roots.

Let me try to factor it: $x^2 - 2x + 2$. Hmm, the discriminant is $4 - 8 = -4$, so the roots are
\[
\frac{2 \pm \sqrt{-4}}{2} = 1 \pm i.
\]
So, it factors as $(x - (1 + i))(x - (1 - i))$. So,
\[
2 - 2x + x^2 = (x - (1 + i))(x - (1 - i)).
\]
So, each term in the product is $(\omega^k - (1 + i))(\omega^k - (1 - i))$.

So, the entire product becomes the product from $k=0$ to $12$ of $(\omega^k - (1 + i))(\omega^k - (1 - i))$. That's the same as the product of $(\omega^k - (1 + i))$ times the product of $(\omega^k - (1 - i))$.

So, I can write it as
\[
\left[\prod_{k=0}^{12} (\omega^k - (1 + i))\right] \cdot \left[\prod_{k=0}^{12} (\omega^k - (1 - i))\right].
\]

Now, I remember that for a root of unity, the product over $(x - \omega^k)$ is equal to $x^{13} - 1$, because the minimal polynomial for $\omega$ is $x^{13} - 1$. So,
\[
\prod_{k=0}^{12} (x - \omega^k) = x^{13} - 1.
\]
So, if I set $x = 1 + i$, then
\[
\prod_{k=0}^{12} (1 + i - \omega^k) = (1 + i)^{13} - 1.
\]
Similarly,
\[
\prod_{k=0}^{12} (1 - i - \omega^k) = (1 - i)^{13} - 1.
\]
So, my product is
\[
[(1 + i)^{13} - 1][(1 - i)^{13} - 1].
\]
So, I need to compute $(1 + i)^{13}$ and $(1 - i)^{13}$.

Hmm, okay, let's compute $(1 + i)^{13}$. I know that $1 + i$ can be written in polar form. The modulus is $\sqrt{1^2 + 1^2} = \sqrt{2}$, and the argument is $\pi/4$.

So,
\[
1 + i = \sqrt{2} \cdot e^{i\pi/4}.
\]
So,
\[
(1 + i)^{13} = (\sqrt{2})^{13} \cdot e^{i13\pi/4}.
\]
Similarly,
\[
1 - i = \sqrt{2} \cdot e^{-i\pi/4},
\]
So
\[
(1 - i)^{13} = (\sqrt{2})^{13} \cdot e^{-i13\pi/4}.
\]
Let me compute $(\sqrt{2})^{13}$. That's $(2^{1/2})^{13} = 2^{13/2} = 2^6 \cdot \sqrt{2} = 64 \sqrt{2}$.

Okay, so
\[
(1 + i)^{13} = 64 \sqrt{2} \cdot e^{i13\pi/4}.
\]
Now, $13\pi/4$ is the same as $3\pi/4 + 2\pi \times 3$, so
\[
e^{i13\pi/4} = e^{i(3\pi/4 + 2\pi \times 3)} = e^{i3\pi/4}.
\]
So,
\[
e^{i3\pi/4} = \cos(3\pi/4) + i \sin(3\pi/4) = -\frac{\sqrt{2}}{2} + i \frac{\sqrt{2}}{2}.
\]
So,
\[
(1 + i)^{13} = 64 \sqrt{2} \left(-\frac{\sqrt{2}}{2} + i \frac{\sqrt{2}}{2}\right) = 64 \cdot (-1) + 64i \cdot 1 = -64 + 64i.
\]
Similarly,
\[
(1 - i)^{13} = 64 \sqrt{2} \cdot e^{-i13\pi/4} = 64 \sqrt{2} \cdot e^{-i3\pi/4} = 64 \sqrt{2} \left(-\frac{\sqrt{2}}{2} - i \frac{\sqrt{2}}{2}\right) = 64 \cdot (-1) - 64i \cdot 1 = -64 - 64i.
\]
So,
\[
(1 + i)^{13} - 1 = (-64 + 64i) - 1 = -65 + 64i.
\]
Similarly,
\[
(1 - i)^{13} - 1 = (-64 - 64i) - 1 = -65 - 64i.
\]
So, the product is $(-65 + 64i) \cdot (-65 - 64i)$. That's the product of a complex number and its conjugate, which is the modulus squared.

So,
\[
(-65)^2 + (64)^2 = 4225 + 4096 = 8321.
\]
So, the product is $8321$.

So, the original product is $8321$.

Now, I need to find the remainder when $8321$ is divided by $1000$. So, $8321$ divided by $1000$ is $8$ with a remainder of $321$. So, the remainder is $321$.

So, the answer is $\boxed{321}$.

Wait, let me double-check my steps. I converted $1 + i$ to polar form, computed its $13$th power, which gave me $-64 + 64i$, then subtracted $1$ to get $-65 + 64i$. Similarly for the other term, I got $-65 - 64i$. Then, multiplying them together, I used the formula $(a + b)(a - b) = a^2 + b^2$, which gave me $65^2 + 64^2$. $65^2$ is $4225$, $64^2$ is $4096$, adding them gives $8321$. Then, $8321 \bmod 1000$ is $321$. So, yeah, that seems correct.

$<$/think$>$ 

The remainder is \boxed{321}.

\end{tcolorbox} 
\end{document}